\documentclass[conference]{IEEEtran}
\IEEEoverridecommandlockouts

\usepackage[linesnumbered,ruled,vlined]{algorithm2e}

\usepackage{cite}
\usepackage{amsmath,amssymb,amsfonts}
\usepackage{algorithmic}
\usepackage{graphicx}
\usepackage{textcomp}
\usepackage{xcolor}

\usepackage{cancel}
\usepackage{CJKutf8}
\usepackage{hyperref}
\usepackage{epstopdf}
\usepackage{enumitem}

\usepackage[T1]{fontenc}
\usepackage{float}
\usepackage[utf8]{inputenc}
\usepackage{multirow} 
\usepackage{makecell}
\usepackage{setspace}
\usepackage{soul}
\usepackage{subcaption}
\usepackage{booktabs}
\usepackage{tikz}

\usepackage{balance}
\definecolor{mylightgreen}{RGB}{144,238,144}
\definecolor{mylightred}{RGB}{255,66,53}
\definecolor{mylightblue}{RGB}{218,227,245} 
\definecolor{darkred}{rgb}{0.70, 0, 0}
\definecolor{darkgreen}{rgb}{0, 0.55, 0}
\definecolor{darkblue}{rgb}{0, 0.0, 0.78}
\definecolor{darkpurple}{rgb}{0.53, 0, 0.50}
\definecolor{purple}{rgb}{0.57, 0.55, 0.78}
\definecolor{iryellow}{rgb}{0.66, 0.82, 0.56}
\definecolor{trolleygrey}{rgb}{0.5, 0.5, 0.49}
\definecolor{tropicalrainforest}{rgb}{1.0, 0.91, 0.7}
\definecolor{glaucous}{rgb}{0.38, 0.51, 0.71}
\definecolor{cardinal}{rgb}{0.7, 0.33, 0.3}
\definecolor{palegreen}{rgb}{0.61, 0.7, 0.35}
\definecolor{pink}{rgb}{0.97, 0.78, 0.65}
\definecolor{orange}{rgb}{0.96, 0.69, 0.51}
\definecolor{purple}{rgb}{0.57, 0.55, 0.78}

\def\BibTeX{{\rm B\kern-.05em{\sc i\kern-.025em b}\kern-.08em
    T\kern-.1667em\lower.7ex\hbox{E}\kern-.125emX}}
\begin{document}


\title{NAT-NL2GQL: A Novel Multi-Agent Framework for Translating Natural Language to Graph Query Language}


\author{
\IEEEauthorblockN{
Yuanyuan Liang{$^\dagger$}, 
Tingyu Xie{$^\ddagger$}, 
Gan Peng{$^\dagger$}, 
Zihao Huang{$^\dagger$}, 
Yunshi Lan{$^\dagger$}, 
Weining Qian{$^\dagger$}
}
\IEEEauthorblockA{
\textit{{$^\dagger$}School of Data Science and Engineering, East China Normal University} \\ 
\textit{{$^\ddagger$}School of Computer Science and Technology, Zhejiang University}
}
leonyuany@stu.ecnu.edu.cn, tingyuxie@zju.edu.cn,{\{gpeng, zhHuang\}}@stu.ecnu.edu.cn,  {\{yslan, wnqian\}}@dase.ecnu.edu.cn
}

\maketitle

\begin{abstract}
       The emergence of Large Language Models (LLMs) has revolutionized many fields, not only traditional natural language processing (NLP) tasks. Recently, research on applying LLMs to the database field has been booming, and as a typical non-relational database, the use of LLMs in graph database research has naturally gained significant attention. Recent efforts have increasingly focused on leveraging LLMs to translate natural language into graph query language (NL2GQL). Although some progress has been made, these methods have clear limitations, such as their reliance on streamlined processes that often overlook the potential of LLMs to autonomously plan and collaborate with other LLMs in tackling complex NL2GQL challenges. To address this gap, we propose NAT-NL2GQL, a novel multi-agent framework for translating natural language to graph query language. Specifically, our framework consists of three synergistic agents: the Preprocessor agent, the Generator agent, and the Refiner agent.
       The Preprocessor agent manages data processing as context, including tasks such as name entity recognition, query rewriting, path linking, and the extraction of query-related schemas. The Generator agent is a fine-tuned LLM trained on NL-GQL data, responsible for generating corresponding GQL statements based on queries and their related schemas. The Refiner agent is tasked with refining the GQL or context using error information obtained from the GQL execution results.
       Given the scarcity of high-quality open-source NL2GQL datasets based on nGQL syntax, we developed StockGQL, a dataset constructed from a financial market graph database, which will be released publicly for future researches. 
       To demonstrate the effectiveness of our proposed NAT-NL2GQL framework, we conducted experiments on the StockGQL and SpCQL datasets. Experimental results reveal that our method significantly outperforms baseline approaches, highlighting its potential for advancing NL2GQL research. The StockGQL dataset can be accessed at:  \href{https://github.com/leonyuancode/StockGQL}{https://github.com/leonyuancode/StockGQL}.
\end{abstract}

\section{Introduction}
\label{sec:intro}
Graph data is gaining prominence in modern data science due to its ability to uncover complex relationships, enhance information connectivity, and facilitate intelligent decision-making. With its distinctive relational representation and efficient processing capabilities, graph data proves invaluable across diverse fields such as finance, healthcare, and social networks, particularly for managing highly connected and structurally complex data~\cite{zhao2022graph,sui2024unleashing}.
Relational databases (DBs) are purpose-built for the efficient storage and management of structured data, while graph data demands specialized graph DBs to enable effective storage and processing~\cite{pavlivs2024graph}. These databases are specifically optimized for accessing and querying graph-structured data, facilitating the efficient manipulation of complex relationships.  
Popular graph databases, including Neo4j, NebulaGraph, and JanusGraph, offer distinct features and applications, empowering users to better analyze and utilize intricate data relationships~\cite{lopes2023scalability,wang2020empirical}.

\begin{figure}[t!]
\centering
\includegraphics[width=0.48\textwidth]{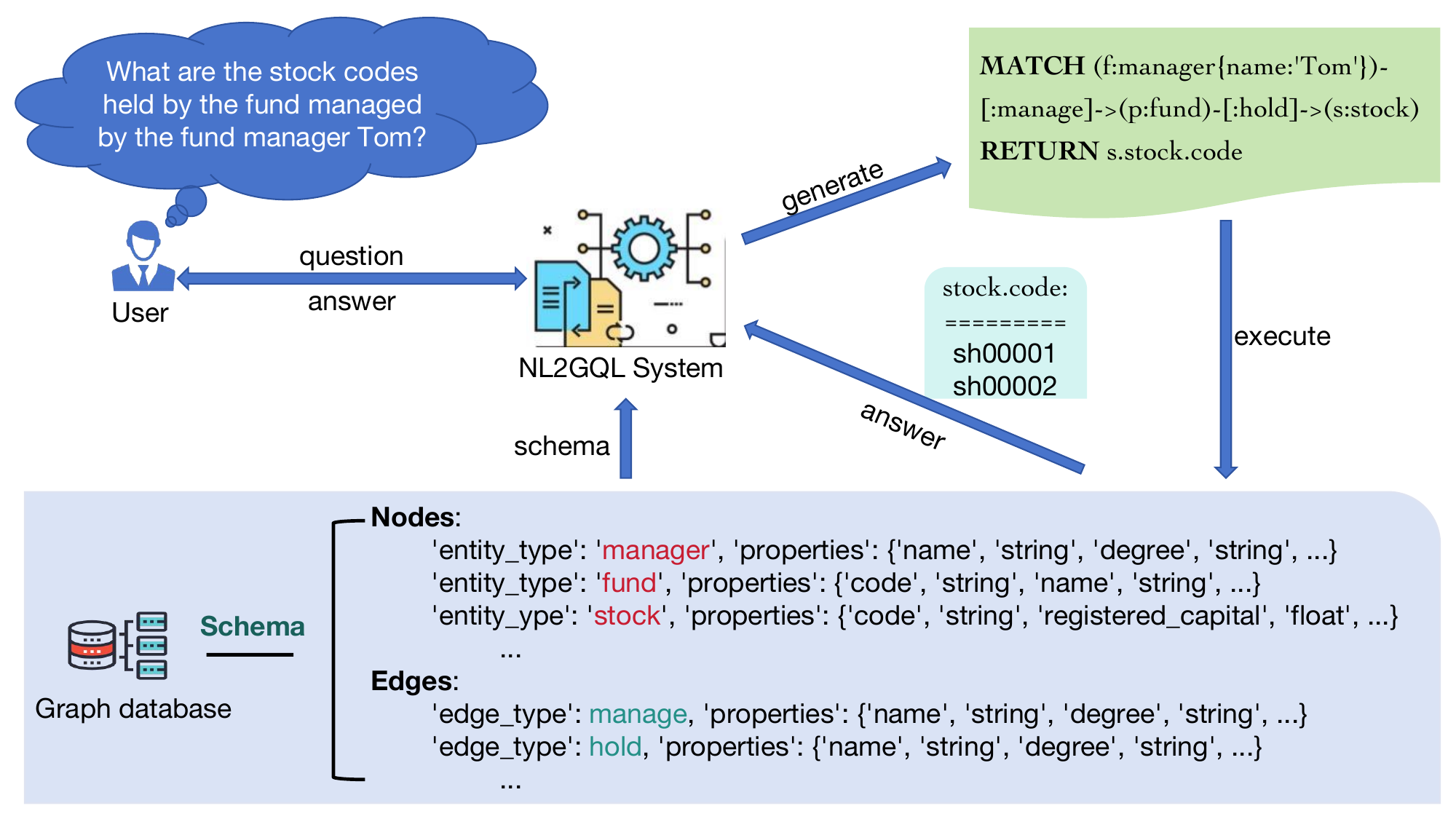}
\caption{The demonstration of the NL2GQL task transforming the user's natural language into a graph query language that can be executed on a graph DB.}
\label{fig:motivation}
\vspace{-1.5em}
\end{figure}

Despite the growing importance of graph data across various domains, ordinary users often encounter significant challenges when interacting with graph DBs due to their specialized and complex operations. The lack of technical expertise among many individuals prevents them from fully harnessing the potential of graph data, thereby limiting its broader adoption in real-world applications~\cite{guo2022spcql}. Furthermore, the intricate and nuanced syntax of graph query languages (GQL) presents additional barriers, particularly for users attempting to translate natural language (NL) into GQL—a task commonly referred to as NL2GQL. These challenges collectively make NL2GQL a highly demanding problem~\cite{liang2024aligning,zhou2024r}.
Figure ~\ref{fig:motivation} illustrates an example of NL2GQL, showcasing its key components: natural language understanding, DB schema comprehension, and the generation and execution of GQL. This highlights the critical need for a system capable of automating the NL2GQL process. By lowering the barriers to entry, such a system would empower users to seamlessly perform data queries and analyses, thus promoting the widespread adoption and practical utilization of graph data.

In many cases, NL2GQL can be viewed as a specialized application of the sequence-to-sequence (Seq2Seq) task. As such, modern approaches have progressed beyond initial template-based methods to focus on generative models for automatically producing GQL. Compared to template-based techniques, generative models offer greater flexibility and accuracy, enabling the handling of more complex query requirements. This advancement effectively meets the evolving demands of application scenarios and rising user expectations.
The study~\cite{guo2022spcql}
was the first to explore the application of a Seq2Seq framework for NL2GQL, utilizing both the Seq2Seq model and copying mechanism to generate GQL.
Although the accuracy of their proposed baseline method is relatively low, just above $2\%$, they also introduced a general-domain NL2GQL dataset, SpCQL.  
Building on the similarities between GQL and SQL, ~\cite{zhao2023cyspider} developed a SQL2Cypher algorithm designed to map SQL queries to Cypher queries. However, this algorithm is limited to constructing an NL2GQL dataset. Furthermore, the substantial differences between GQL and SQL present significant challenges in transferring Text2SQL methods to NL2GQL.
The study \cite{tran2024robust} proposes the CoBGT model, which combines BERT, GraphSAGE\cite{hamilton2017inductive}, and Transformer, to perform key-value extraction, relation-property prediction, and Cypher query generation.
The work ~\cite{liang2024kei} introduces the KEI-CQL framework, which utilizes pre-trained language models to extract semantic features from natural language queries and populate predefined slots in Cypher query templates, effectively addressing the NL2GQL challenge.

The advent of large language models (LLMs) has recently revolutionized performance across various natural language processing (NLP) tasks. With their exceptional generalization and generation capabilities, LLMs have found widespread applications across diverse domains. Notably, their integration into database research has gained significant traction in recent years~\cite{zhu2024autotqa,ren2024purple,peng2024online,zhou2023d,lao2023gptuner}, driven by their potential to bridge the gap between NL and structured query languages, enabling more intuitive and efficient database interactions.
Recently, there has been growing research interest in applying LLMs to graph databases, with a primary focus on the NL2GQL task.
This study ~\cite{tao2024finqa} attempts to address this task through heuristic prediction and LLM-based revision. Although the method is relatively simple, it has demonstrated some effectiveness in specific domains.
This work ~\cite{zhou2024r} leverages the interpretative strengths of smaller models during the initial ranking and rewriting stages, while capitalizing on the superior generalization and query generation capabilities of larger models for the final transformation of NL into GQL formats.
This paper ~\cite{liang2024aligning} proposes a method for constructing an NL2GQL dataset based on a domain-specific graph database and recommends using tokenization to extract the related schema and incorporate it into the context to enhance the model's accuracy. \textbf{LLM-based methods exhibit some effectiveness in solving NL2GQL tasks, but their streamlined approach carries a major challenge—error accumulation.} If the related schema is extracted incorrectly, it significantly increases the likelihood of generating flawed GQL. For instance, as shown in Figure ~\ref{fig:motivation}, the correct related schema for the query includes the nodes “manager,” “fund,” and “stock,” along with the edges “manage” and “hold.” However, if an incorrect related schema is extracted that only includes “manager,” “fund,” and “manage,” the generated GQL might be: 
\textit{MATCH (f:manager\{name:'Tom'\})-[:manage]->(p:fund) RETURN s.fund.code}, which would undoubtedly produce incorrect results.

In this study, we aim to address this challenge by designing a multi-agent framework, \textbf{NAT-NL2GQL},a \textbf{N}ovel Multi-\textbf{A}gent Framework
for \textbf{T}ranslating \textbf{NL2GQL}—as illustrated in Figure ~\ref{fig:Overview}. The framework comprises three synergistic agents:  \textit{Preprocessor} agent, \textit{ Generator} agent, and \textit{Refiner} agent.
The Preprocessor agent focuses on data preprocessing tasks, such as extracting values from the graph DB, performing named entity recognition (NER), rewriting user queries, linking paths, and extracting related schemas as the context. 
The Generator agent generates the corresponding GQL based on the provided context and user queries. It is a fine-tuned LLM trained on the NL-GQL dataset. 
Finally, the Refiner agent refines the GQL or context via the error information obtained from the execution results of the GQL.
The working flow could return to Preprocessor agent and Generator agent with informative interactions.
This entire process operates iteratively, with a maximum of three iterations. 
Additionally, due to the lack of high-quality open-source NL2GQL datasets based on nGQL syntax, we have developed StockGQL, a dataset derived from a financial market NebulaGraph DB.
To evaluate the performance of our framework, we conducted a comprehensive assessment using the StockGQL and SpCQL datasets. The results show that our framework significantly outperforms baseline methods in improving the accuracy of NL2GQL tasks. Additionally, ablation experiments confirm that each module within our framework plays a crucial role in enhancing task accuracy.

\textbf{Key Contributions.} To summarize, this paper makes the following contributions:
\begin{itemize}
    \item 
First, to alleviate error accumulation inherent in streamlined methods, we designed a collaborative and iterative multi-agent framework to tackle the NL2GQL task. 
    \item 
Second, we constructed the StockGQL dataset, which could perform as a testbed for the domain-specific NL2GQL task.
    \item
Third, our proposed method surpassed the baseline methods on both StockGQL and SpCQL datasets, which denotes the new state-of-the-art NL2GQL results in both general and specific domains.
\end{itemize}


\textbf{Organization.} The rest of the paper is organized as follows: Section 2 introduces the preliminaries. Section 3 covers the construction of the StockGQL dataset and discusses the overall architecture of NAT-NL2GQL. Section 4 delves into the detailed implementation of NAT-NL2GQL. Section 5 presents the experimental evaluation. Related work is reviewed in Section 6, and the paper concludes in Section 7.

\begin{figure*}[ht]
\centering
\includegraphics[width=0.98\textwidth]{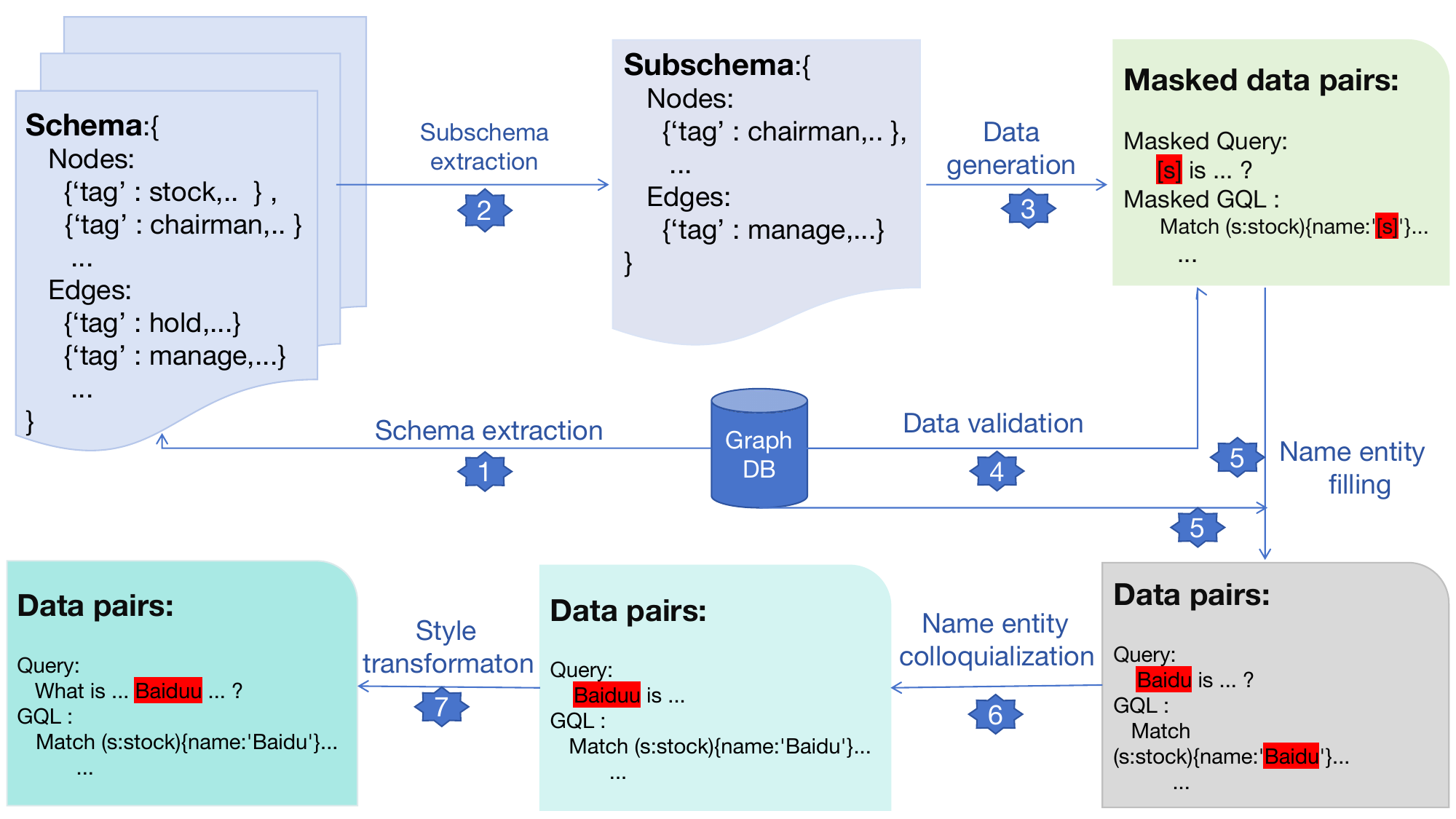}
\caption{Our StockGQL dataset construction flowchart highlights the areas of improvement in our method, marked in red.}
\label{fig:data_build}
\end{figure*}

\section{PRELIMINARIES}

\noindent \textbf{NL2GQL Task Definition}. We formally define a NL2GQL task. 
The input consists of a NL query $\mathcal{X}$ and a graph DB $\mathcal{G}$, which is represented as $\mathcal{G} = {(s, r, o) \mid s, o \in \mathcal{E}, r \in \mathcal{V}}$. Here, $\mathcal{E}$ and $\mathcal{V}$ denote the sets of vertices and edges, respectively. The objective is to generate a correct GQL query $\mathcal{Q}$. 
Hence a NL2GQL system can be formulated as:
$$\mathcal{\hat{Q}}=NL2GQL(\mathcal{X}, \mathcal{G}).$$
Here $\mathcal{\hat{Q}}$ is the GQL prediction.

\noindent \textbf{LLM-based NL2GQL Systems}. 
Multiple LLM-based methods have been proposed to address the NL2GQL task. These methods typically follow a unified paradigm that incorporates techniques such as in-context learning. In-context learning, which is widely adopted in current models, enables LLMs to generate correct answers by including a few examples directly within the prompt. This approach can be formalized as follows:
$$\mathcal{\hat{Q}} = LLM_{ICL}(\mathcal{I}, \mathcal{E},\mathcal{X})$$
Here, \(\mathcal{I}\) represents the task description, \(\mathcal{E}\) denotes the demonstrations derived from annotated datasets, and \(\mathcal{X}\) is the input question, which can optionally be empty.

\section{StockGQL Dataset Build}

The lack of high-quality open-source NL2GQL datasets has hindered progress in this research area. While several datasets based on Neo4j's Cypher syntax are available, NebulaGraph—a widely used, efficient, enterprise-grade open-source distributed graph database—remains without datasets specifically designed for its nGQL syntax. To bridge this gap and advance future research, we follow the approach in \cite{liang2024aligning}, which utilizes the self-instruct \cite{wang2022self} method, to develop the StockGQL dataset based on a real-world financial stock scenario graph DB, which we crawled from stock trading websites and applied privacy processing to some named entities and real data. As shown in Figure ~\ref{fig:data_build}, our method consists of several components.

\noindent \textbf{Schema Extraction.} As Step 1 of Figure ~\ref{fig:data_build} illustrates, we extract the schema from the graph DB, which includes identifying the nodes, edges, and their attributes. This step forms the foundation for constructing a structured representation of the graph, enabling further processing and query generation in subsequent steps.
\noindent \textbf{Subschema Extraction.} A subschema is the schema of a subgraph derived from the complete graph, containing only partial information from the graph's overall schema. Step 2 of Figure ~\ref{fig:data_build} is to extract a subschema from the schema. We apply specific rules to identify all possible path combinations, ranging from 0-hop to 6-hop paths, to address the limitation of the study ~\cite{liang2024aligning}, which does not guarantee that the dataset covers all possible link combinations. Our approach ensures a comprehensive coverage of potential relationships within the graph, enabling a more robust dataset for training and evaluation.

\noindent \textbf{Data Generation.} Step 3 in Figure ~\ref{fig:data_build} illustrates the data generation module. The generation process algorithm is illustrated in Algorithm ~\ref{algoritm:data_generate}. We employ the ICL method, continuously sampling $K$ data points randomly from the data pool to construct masked NL-GQL data pairs, each of which consists of a \textit{masked query} and a \textit{masked GQL}..
These generated masked NL-GQL data pairs are then returned to the pool. 
The masked query and masked GQL represent the masking of entity names in both the query and the corresponding GQL. An example is shown below:
\begin{itemize}[label={}, labelsep=2pt, leftmargin=15pt]
\item {\textit{Masked query : What is the code of stock [s]?}}
\item {
\textit{Masked GQL : MATCH (s:stock\{name:'[s]'\})}
\begin{itemize}[label={}, labelsep=2pt, leftmargin=65pt]
\item \textit{RETURN s.stock.code}
\end{itemize}
}
\end{itemize}
We use the placeholder \texttt{[s]} to represent stock entity names in both the natural language query and the corresponding GQL.

We generate each subschema for $m$ times to cover as many attributes of all entities as possible. Using the self-instruct approach, the process iterates until all subschemas have been covered, at which point it will terminate.

\begin{algorithm}[]
\small
\caption{Masked NL-GQL Data Pairs Generation}
\label{alg:nlgql-data-pair}
\KwIn{A set of subschemas; Data pool; Number of demonstrations $k$;  Iterations number $m$; Task description $I$; Sample size $k$; ICL formula $LLM_{ICL}$}
\KwOut{Updated Data pool with masked NL-GQL data pairs}
\ForEach{$subschema$ in subschemas}{
    \For{$i = 1$ \KwTo $m$}{
        Sample $k$ items from Data pool\;
        Build demonstrations \(\mathcal{E}\) using the sampled data\;
        Generate Masked NL-GQL Data Pairs\;
        \Indp
         $d\_list \leftarrow$  $LLM_{ICL}$($I$,\(\mathcal{E}\), $subschema$)\;
         \Indm
        Add $d\_list$ to Data pool\;
    }
}
 \label{algoritm:data_generate}
\end{algorithm}

\noindent \textbf{Data Validation.} This step filters out erroneous data where $NL$ and $GQL$ are inconsistent. We follow the approach outlined in ~\cite{liang2024aligning}, using an entity-filled, CoT-based GQL2NL method to generate $NL'$ from $GQL$. The data is then filtered based on low embedding similarity between $NL$ and $NL'$. As a result, we obtain a large number of high-quality masked NL-GQL data pairs.

\noindent \textbf{Name Entity Filling.} This process helps in creating a diverse and representative dataset by leveraging the rich structure and variety of the graph data. It not only improves the alignment between natural language questions and their corresponding GQL queries but also aids in training and evaluating models on tasks that require precise understanding and execution of graph queries. This step is crucial for bridging the gap between theoretical schema-based generation and practical, real-world graph operations. After filling and executing the GQL, the answer for this data point can be obtained.

\noindent \textbf{Name Entity Colloquialization.} In this step, we randomly select a dataset containing named entities from the constructed data and manually rewrite the named entities in both the NL and GQL to their abbreviated forms. This process simulates real-world scenarios where users often use shortened versions of entity names, ensuring the dataset reflects a variety of linguistic expressions. By incorporating abbreviations, the dataset becomes more robust, enabling models to better handle variations in how entities are referenced, thereby enhancing their flexibility and generalization in practical applications.

\noindent \textbf{Style Transformation.} The final step is to transform the style of the NL by adjusting the wording and phrasing to ensure that it fits different contexts or user needs while preserving the core meaning of the original NL. This includes rephrasing the NL to align with various linguistic preferences, tone, or formality levels, ensuring it remains clear and relevant across different scenarios without altering its intended purpose.

As a result, we form StockGQL dataset.
Each data entry in the dataset consists of a Qid, Query\_masked along with its corresponding GQL\_masked, a Query with its corresponding GQL, and a SubSchema containing Nodes and Edges. It also includes Query, Masked\_name, and Oral\_name, as well as the corresponding Answer. An example from the Stock dataset is as follows:

\begin{itemize}
    \item \textit{\textbf{Qid}}: 
    {\small 10}
    \item \textit{\textbf{Query\_masked}}: \\
    {\small \begin{CJK*}{UTF8}{gbsn} [c]是董事长的股票关联的产业下游的产业有哪些？\end{CJK*} \\
    \textit{(What are the downstream industries related to the industries associated with the chairman [c]'s stock?)}}

    \item \textit{\textbf{GQL\_masked}}: \\
    {\small MATCH (c:chairman\{name:'[c]'\})\\
    -[:is\_chairman\_of]->(s:stock)-[:associate]->(i1:industry)-[:affect]->(i2:industry) RETURN i2.industry.name}

    \item \textit{\textbf{Query}}: \\
    {\small \begin{CJK*}{UTF8}{gbsn} 梁dong是董事长的股票关联的产业下游的产业有哪些？\end{CJK*} \\
     \textit{(What are the downstream industries related to the industries associated with the chairman Liang Dong's stock?)}}

    \item \textit{\textbf{GQL}}: \\ 
    {\small MATCH (c:chairman\{name:'\begin{CJK*}{UTF8}{gbsn}梁东\end{CJK*}'\})\\
    -[:is\_chairman\_of]->(s:stock)-[:associate]->(i1:industry)-[:affect]->(i2:industry) RETURN i2.industry.name}

    \item \textit{\textbf{SubSchema}}: \\
    {\small nodes : ["chairman", "stock",, "industry"]}\\
    {\small edges : ["is\_chairman\_of", "associate", "affect"]}

    \item \textit{\textbf{Masked\_name}}: \\
    {\small [c] : \begin{CJK*}{UTF8}{gbsn}梁东\end{CJK*}'}

    \item \textit{\textbf{Oral\_name}}: \\
    {\small \begin{CJK*}{UTF8}{gbsn}梁dong\end{CJK*}': \begin{CJK*}{UTF8}{gbsn}梁东\end{CJK*}'\}}

     \item \textit{\textbf{Answer}}: \\
    {\small i2.industry.name : ["\begin{CJK*}{UTF8}{gbsn}电脑硬件\end{CJK*}(Computer Hardware)", "\begin{CJK*}{UTF8}{gbsn}汽车\end{CJK*}(Car)",
    "\begin{CJK*}{UTF8}{gbsn}金融服务\end{CJK*}(Financial services)"]}\\
    
\end{itemize}

After constructing the data, we conducted a statistical analysis, as shown in Table~\ref{dataset_hops}. The number of hops in a query reflects the complexity of the problem: the more hops, the greater the complexity. As seen in the table, 63\% of the data involve more than 2 hops, while 26\% involve more than 3 hops. Additionally, it contains 12 types of nodes, 13 types of edges, and 62 types of properties. Overall, StockGQL is an NL2GQL dataset based on the nGQL syntax, designed to handle complex multi-hop, multi-type queries. We hope that the open-source release of this dataset will contribute to future research in the field and promote the development of more advanced NL2GQL models.

\begin{table}[ht!]
\centering
\small
\resizebox{0.49\textwidth}{!}{
\begin{tabular}{lllllll}

\toprule
\textbf{Dataset} & \textbf{0-hop} & \textbf{1-hop} & \textbf{2-hop} & \textbf{3-hop} & \textbf{4-hop} & \textbf{Others} \\
\midrule
Train (4572)  &  528    &  1167     &   1666    &  942  &  253    &  16      \\
Dev (655)     &  70     &  163      &   207     &  155   &  54     &  6      \\
Test (1229)   &  172    &  378      &   414     &  194   &  62    &  9     \\
\bottomrule
\end{tabular}
}
\caption{\label{dataset_hops}
Statistics on the number of hops contained in StockGQL dataset.
}
\end{table}

\begin{figure*}[ht]
\centering
\includegraphics[width=0.98\textwidth]{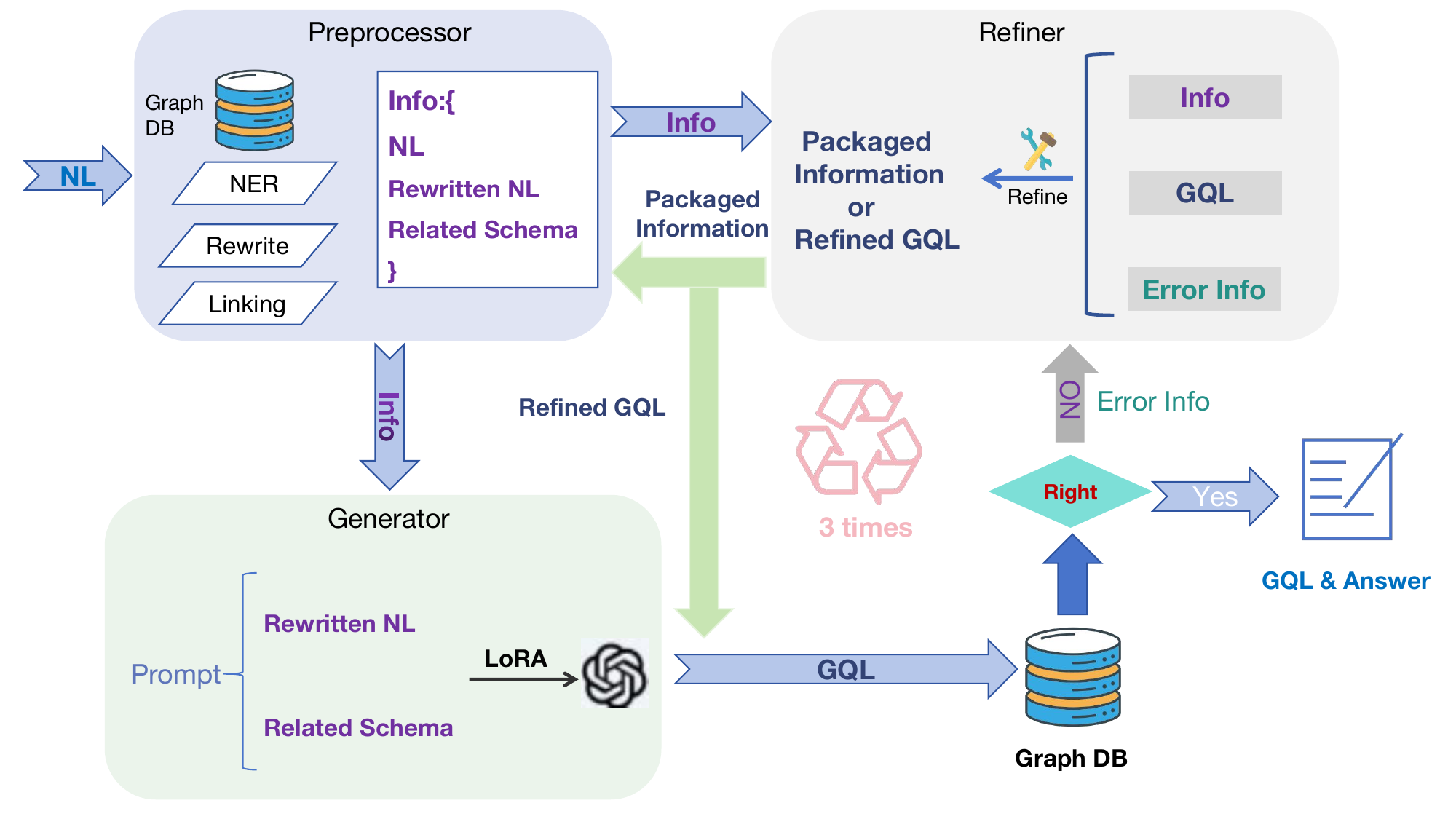}
\caption{Our NAT-NL2GQL framework consists of three synergistic agents: the Preprocessor agent, the Generator agent, and the Refiner agent. The entire process follows a cyclic and iterative flow, with the three agents collaboratively handling data preprocessing, GQL generation, and GQL refinement. 
}
\label{fig:Overview}
\end{figure*}

\section{Method}
In this section, we provide a detailed explanation of the workflow of NAT-NL2GQL and the roles of each component. As illustrated in Figure~\ref{fig:Overview}, NAT-NL2GQL consists of three synergistic agents: the \textit{Preprocessor} agent, the \textit{Generator} agent, and the \textit{Refiner} agent. The Preprocessor agent is primarily responsible for data preprocessing, extracting auxiliary information such as the related schema and written query to facilitate GQL generation. The Generator agent generates the GQL based on the provided context. The Refiner agent rewrites the GQL directly or modifies the context content based on error messages from GQL execution. The specific data flow is as follows: The results from the Preprocessor agent are sent to both the Generator and Refiner agents. Error messages from the execution results of the GQL generated by the Generator agent are sent to the Refiner agent. Based on the received information, the Refiner agent determines whether to modify the GQL directly or assemble the information and send it back to the Preprocessor agent. The three agents collaborate seamlessly to perform the task.

In the following sections, we will provide a detailed description of each agent: the Preprocessor Agent in Section \ref{preprocessor}, the Generator Agent in Section \ref{generator}, and the Refiner Agent in Section \ref{refine}.

\subsection{Preprocessor Agent}
\label{preprocessor}
As highlighted in~\cite{liang2024aligning,zhou2024r}, extracting the NL-relevant schema from the complete graph DB schema provides three significant benefits: first, it reduces the schema size, helping to avoid context length limitations in the model; second, it eliminates irrelevant schema noise, thereby enhancing the accuracy of the generated GQL; and third, leveraging the relevant schema instead of the full schema substantially decreases the time required for LLMs to generate GQL.
The primary function of the Preprocessor agent is to extract schemas relevant to NL by utilizing both the schema information from the graph DBs and the associated data values. Additionally, it aligns named entities in the query with those in the graph DBs and rewrites the query as needed. The Preprocessor agent includes LLM-based NER, entity alignment,  related schema revise, linking completion and question rewriting.

\begin{figure}[ht]
\centering
\resizebox{0.49\textwidth}{!}{
  \begin{tikzpicture}
    \node[draw, rectangle, inner sep=2mm, fill=mylightblue] (rect) {
      \begin{minipage}{\linewidth}
      \textbf{Instruction:}\\
You are an expert in the NLP field. I would appreciate your assistance with an NER task. Given entity label set: {label set}. Refer to the given example. Based on the provided entity label set, please recognize the named entities in the given Question. Please directly output the answer.
\\

\textbf{Output Format: }\\
In JSON format, for example: \{Entity Name: Entity Type, Entity Name: Entity Type\}.\\

\textbf{Here are some examples:}\\
\{EXAMPLES\}\\
======== \textbf{Predict} ========\\
\textbf{Question:}\\
\{QUESTION\}\\
\textbf{Answer:}\\
      \end{minipage}
    };
  \end{tikzpicture}
}
\caption{Prompt for performing Name Entity Recognition on questions using ChatGPT-4o.}
\label{fig:ner}
\end{figure}

\noindent \textbf{LLM-based NER}. Extracting named entities from NL is essential for identifying the related schema. Previous studies have demonstrated that LLMs can effectively recognize named entities within sentences~\cite{xie2023empirical,xiao2024chinese,xu2023large}. Building on this capability, the Preprocessor agent utilizes LLM-based NER to extract entities from the query, which are then employed to pinpoint the most relevant parts of the schema. This process is pivotal for reducing the schema search space and establishing precise mappings between query entities and their counterparts in the graph DB, ensuring that subsequent GQL generation is accurate and contextually appropriate. To achieve this, we employ ChatGPT-4o for name entity extraction, following the prompt structure depicted in Figure~\ref{fig:ner}.


\begin{figure}[!htbp]
\centering
\resizebox{0.49\textwidth}{!}{
  \begin{tikzpicture}
    \node[draw, rectangle, inner sep=2mm, fill=mylightblue] (rect) {
      \begin{minipage}{\linewidth}
      \textbf{Instruction:}\\
You are an expert in the NLP field. I am working on an information extraction task that involves identifying the related schema potentially relevant to a given question from a graph DB schema. I have already extracted the Candidate Related Schema. Please assist me in verifying whether the Candidate Related Schema contains any redundancies and ensure that each one is necessary. Based on the provided examples, kindly provide the correct Candidate Related Schema. \\

\textbf{Candidate Related Schema:}\\
-The complete Schema structure of Candidate Related Nodes and Candidate Related Edges.\\

\textbf{Output Format:} \\
Please follow the format in the Examples. Directly output the result you consider correct after "Related Schema:" .\\

\textbf{Here are some examples:}\\
\{EXAMPLES\}\\
======== \textbf{Predict} ========\\
\textbf{Question:}\\
\{QUESTION\}\\
\textbf{Candidate Related Schema:}\\
\{Candidate\_related\_schema\}\\

\textbf{Related Schema:}\\
      \end{minipage}
    };
  \end{tikzpicture}
}
\caption{Prompt for revising the related schema.}
\label{fig:filter}
\end{figure}

\noindent \textbf{Entity Alignment}. After extracting named entities from natural language, the next step is to align these entities with the corresponding entity names in the graph DB. This alignment ensures that the extracted entities are accurately mapped to the relevant nodes or edges in the DB, maintaining consistency and enabling precise query generation across both the natural language and graph representations. Our approach begins by extracting named entities from each entity in the graph DB to build a dictionary  \(\mathcal{D}\), where each key is an entity type, and each value is a list of names for entities of that type. Next, we compare the name of each entity extracted in the initial step with the names in this dictionary. If an exact match is found, the corresponding entity type name is assigned. For entities without an exact match, we apply locality-sensitive hashing (LSH)~\cite{datar2004locality} to select the name of the most similar entity. This process can be formulated as:
$$\mathcal{\hat{D}}=LSH(\mathcal{Z}, \mathcal{D},\mathcal{\gamma} )$$
$$\hat{d} = \arg\max_{d_i \in \mathcal{\hat{D}}} \text{Cosine}(Emb(\mathcal{X}), Emb(d_i))$$
Here, \(\mathcal{Z}\) denotes the extracted name entity from the NL via the LLM-based NER, \(\mathcal{D}\) represents the entire entities dictionary extracted from the graph DB, and \(\mathcal{\hat{D}}\) represents the entities extracted with the LSH similarity to \(\mathcal{X}\) according to a threshold \(\mathcal{\gamma}\).
\textit{Emb}(\(\mathcal{X}\)) represents the embedding of  \(\mathcal{X}\) encoded via all-MiniLM-L12-v1.
\(\hat{d}\)  represents the entity names extracted based on the cosine similarity to \(\mathcal{X}\).
Once the entity alignment is completed, we will obtain the entity names along with their corresponding entity types.

\begin{algorithm}[]
\caption{Linking Completion Algorithm}
\KwIn{Graph Schema \( G = (V, E) \); Identified Entities \( E_{\text{identified}} \); Identified Edges \( R_{\text{identified}} \)}
\KwOut{Connected Subgraph \( SG = (V_{\text{subgraph}}, E_{\text{subgraph}}) \)}

\SetKwFunction{FMain}{LinkCompletion}
\SetKwProg{Fn}{Function}{:}{}
\Fn{\FMain{$G, E_{\text{identified}}, R_{\text{identified}}$}}{
    \( V_{\text{subgraph}} \gets \emptyset \) \\
    \( E_{\text{subgraph}} \gets \emptyset \) \\
    
    \ForEach{entity \( v_i \in E_{\text{identified}} \)}{
        \( V_{\text{subgraph}} \gets V_{\text{subgraph}} \cup \{v_i\} \) \\
    }
    
    \ForEach{edge \( r_j \in R_{\text{identified}} \)}{
        \( E_{\text{subgraph}} \gets E_{\text{subgraph}} \cup \{r_j\} \) \\
    }
    
    \ForEach{edge \( e_k \in E_{\text{subgraph}} \)}{
        \ForEach{neighbor \( v_l \in \text{neighbors}(e_k) \)}{
            \( V_{\text{subgraph}} \gets V_{\text{subgraph}} \cup \{v_l\} \) \\
            \( E_{\text{subgraph}} \gets E_{\text{subgraph}} \cup \{e_k\} \) \\
        }
    }
    
    \While{\( V_{\text{subgraph}} \) is not connected}{
        Find the minimum edge to add that connects two disconnected components \\
        \( E_{\text{subgraph}} \gets E_{\text{subgraph}} \cup \{ \text{min edge} \} \) \\
    }
    
    \Return \( SG = (V_{\text{subgraph}}, E_{\text{subgraph}}) \)
}
\label{linkCompletion}
\end{algorithm}

\noindent \textbf{Linking Completion}. Even though we obtain many entity types, these entities may not necessarily form a connected subgraph. Moreover, to address questions that require reasoning across multiple entity types—where not all types are explicitly referenced—it is essential to complete the links with related entities~\cite{liang2024aligning}. Additionally, to gather as much related schema information as possible, we begin by extracting entity names and attribute names from the graph DB schema and then check for matches in the natural language query. If a match is found, we retrieve the corresponding entity type. We then filter out duplicate entity types and identify the edges connecting the entities. After this, we will have an entity list and an edges list. 
Finally, we use the algorithm shown in ~\ref{linkCompletion} to complete any intermediate entities and relationships, ultimately producing a candidate related schema.

\noindent \textbf{Related Schema Revision}. In the previous step, we obtained the candidate related schema. However, due to the use of a relaxed rule when selecting entity types, the extracted related schema may contain redundant nodes and edges. To address this, we apply further filtering techniques to remove these redundancies, ensuring that only the most relevant and essential entities and relationships remain. To accomplish this, we use ChatGPT-4o for the filtering step, with the specific prompt shown in Figure ~\ref{fig:filter}. Subsequent experimental results also have demonstrated that this step significantly enhances the accuracy of the extracted Related Schema.

\noindent \textbf{Question Rewriting}. Since the query may include colloquial terms or abbreviations, these need to be aligned with the graph DB entities and the query rewritten for accurate GQL generation. After aligning the named entities with those in the graph DB, any mismatches must be replaced with the corresponding entity names. During the entity alignment step, many entities may be extracted that do not match exactly, but the subsequent related schema revision step will filter out some of these irrelevant entities. This process requires rewriting the original NL entity names to ensure consistency with the graph DB.
For example,the original query : 

\begin{itemize}[label={}, labelsep=2pt, leftmargin=15pt]
\item {
\textit{
{\small \begin{CJK*}{UTF8}{gbsn}{\color{mylightred}梁dong}是董事长的股票关联的产业下游的产业有哪些？\end{CJK*} \\
    \textit{(What are the downstream industries related to the industries associated with the chairman Liang Dong's stock?)}}
}}
\end{itemize}

can be revised to :
\begin{itemize}[label={}, labelsep=2pt, leftmargin=15pt]
\item {
\textit{
{\small \begin{CJK*}{UTF8}{gbsn}{\color{mylightred}梁东}是董事长的股票关联的产业下游的产业有哪些？\end{CJK*} }
}}
\end{itemize}

\subsection{Generator Agent}
\label{generator}
Once the data pre-processing is complete, the next step is to generate the GQL based on the information obtained. Parameter Efficient Fine-Tuning (PEFT) techniques have been shown to significantly reduce the number of fine-tuned parameters and optimize memory usage while maintaining performance~\cite{ding2023parameter,fu2023effectiveness,dettmers2024qlora,liu2022few}. As one of the most widely recognized methods in PEFT, LoRA~\cite{hu2021lora} has found broad application in various fields, including natural language processing and graph query generation. By fine-tuning only a small subset of parameters, LoRA enables efficient adaptation of large pre-trained models, making it particularly effective for tasks like GQL generation, where model size and computational efficiency are critical. 
Therefore, we choose LoRA for fine-tuning the selected base LLMs. As shown at the lower left corner of Figure ~\ref{fig:Overview}, we concatenate the original NL, the rewritten NL, and the related Schema to create a prompt for fine-tuning the LLMs. 
It is worth noting that during training, we use the golden related schema extracted from the labeled GQL, while during inference, the related schema is predicted by the Preprocessor agent. Additionally, the Generator agent is not restricted to fine-tuning LLMs; it can also fine-tune smaller models or directly leverage closed-source APIs, such as ChatGPT-4o.

\subsection{Refiner Agent}
\label{refine}
Many studies have demonstrated that rewriting SQL queries with syntax errors can improve the accuracy of Text2SQL~\cite{pourreza2024chase,wang2023mac,talaei2024chess}. However, these methods typically rely on LLMs to rewrite queries with syntax errors. Through experimental validation, we have found that such rewrites often involve only minor modifications to the original query, which may not be sufficient to address more complex issues. Moreover, the error information typically highlights only the first error encountered during execution, making it unsuitable for handling queries with multiple errors. Most importantly, if the related schema or written query from earlier steps is incorrect, correcting the GQL syntax alone may not resolve the issue, as it may still not align with the original query. In such cases, the error information should be used to revisit the auxiliary information extracted in earlier steps. \textbf{Our approach differs by using the error information to determine whether to directly rewrite the GQL or package the information to the Preprocessor agent, treating both the GQL and error details as historical data for re-execution.}

\begin{figure}[]
\centering
\includegraphics[width=0.48\textwidth]{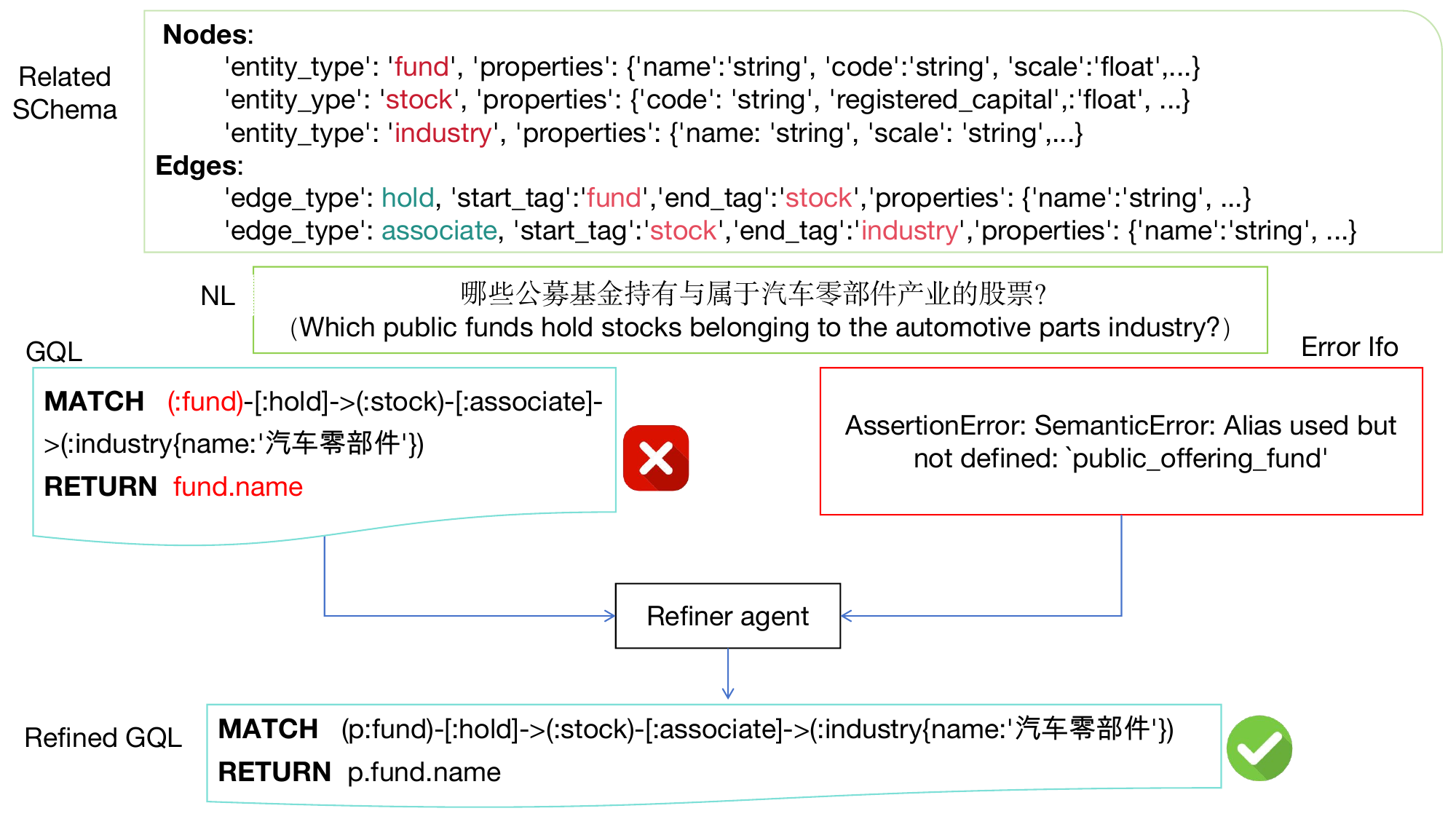}
\caption{A Refiner example: From the example, we can see that the Refiner agent corrected the GQL based on the error information.  
}
\vspace{-1.5em}
\label{fig:example}
\end{figure}

It is important to note that even after modifying the GQL, it may still contain errors. Therefore, we set a limit on the number of iterations in the process, and once the limit is reached, the process will terminate even if the GQL is still incorrect. The refine prompt is shown in Figure ~\ref{fig:refiner}. We enable the Refiner agent to decide, based on the error information, whether to modify the GQL directly or save the historical data to restart the entire process. Examples are shown in Figure~\ref{fig:example}. 
As shown in the figure, the Refiner agent directly corrected the syntax error in the generated GQL.

So far, we have covered all aspects of NAT-NL2GQL. It is important to note that the three agents in NAT-NL2GQL can be combined in various configurations, such as using only the Generator and Refiner agents, or the Preprocessor and Generator agents. However, the corresponding prompts would need to be adjusted accordingly. Furthermore, the base LLM for each agent can be chosen and replaced based on the specific needs of the task.


\begin{figure}[ht!]
\centering
\resizebox{0.49\textwidth}{!}{
  \begin{tikzpicture}
    \node[draw, rectangle, inner sep=2mm, fill=mylightblue] (rect) {
      \begin{minipage}{\linewidth}
      \textbf{Instruction:}\\
You are an expert in NebulaGraph DBs, with specialized expertise in nGQL. A prior attempt to execute a query 
did not produce the expected results, either due to execution errors or because the returned output was empty or incorrect.
Your task is to analyze the issue using the provided related schema of query and the details of the failed execution. 
Based on this analysis, you should offer a corrected version of the nGQL. Ensure adherence to the nGQL conventions
for naming variables, entities, and attributes (e.g., `s.stock.name`) and verify that all conditional filters
use `==` syntax, such as `s.stock.name == '[s]'`.\\

\textbf{Procedure:}\\
1. Analyze Query Requirements:\\
- Question: Consider what information the query is supposed to retrieve.\\
- nGQL: Review the nGQL query that was previously executed and led to an error or incorrect result.\\
- Error: Analyze the outcome of the executed query to identify why it failed (e.g., AssertionError).\\
2. Determine whether the Related Schema is correct.\\
- Based on the above information, first determine whether the extracted related schema is correct.\\
- If related schema is not correct, directly output "Info Error". Otherwise, modify the nGQL query to address the identified issues, ensuring it correctly fetches 
the requested data according to the graph DB schema and query requirements.\\

\textbf{Output Format:}\\
Based on whether the determined Related Schema is correct, output either "Info Error" or your corrected query.
The corrected query as a single line of nGQl code. Ensure there are no line breaks within the query.\\

\textbf{Here are some examples:}\\
\{EXAMPLES\}\\
======== \textbf{Predict} ========\\
\textbf{Question:}\\
\{QUESTION\}\\
\textbf{Related Schema:} \\
\{RELATED\_SCHEMA\}\\
\textbf{nGQL:}\\
\{nGQL\}\\
\textbf{Error:}\\
\{ERROR\}\\
\textbf{Output:}
      \end{minipage}
    };
  \end{tikzpicture}
}
\caption{The prompt template used for GQL refine.}
\label{fig:refiner}
\end{figure}

\begin{table*}[ht!]
\centering
\small
\tabcolsep=1.1em
\renewcommand\arraystretch{0.8}
\begin{tabular}{l l c c c c }
\toprule
\multirow{2}{*}{\textbf{Method}} & \multirow{2}{*}{\textbf{Backbones}}& \multicolumn{2}{c}{\textbf{StockGQL}} & \multicolumn{2}{c}{\textbf{SpCQL}}\\
 \cmidrule{3-4} \cmidrule{5-6}
& & \textbf{EM(\%)} & \textbf{EX(\%)} & \textbf{EM(\%)} & \textbf{EX(\%)} \\
\midrule

\multirow{4}{*}{ICL(K=4)}
                              & GLM-4-9B-Chat       & 30.43  & 28.23 & 7.03 & 8.22  \\
                              & Qwen2.5-14B-Instruct & 31.16   & 27.18 & 7.87  & 8.92 \\
                              & LLaMA-3.1-8B-Instruct &  24.74  & 23.35 & 7.42  & 8.21 \\
                              & LLaMA-3.2-3B-Instruct & 8.62   & 8.46 & 6.03  & 7.27 \\
                              & ChatGPT-3.5-Turbo    & 29.29 & 29.21  & 7.37 & 7.62 \\
                              & ChatGPT-4o   & 40.68  & 38.08  &  9.22  & 10.26 \\
\midrule
\multirow{2}{*}{Fine-Tuning(full schema)}  &GLM-4-9B-Chat   & 80.96  & 81.77  & 53.86  &   52.12   \\
                            & Qwen2.5-14B-Instruct & 82.67 & 83.65  & 53.91  & 51.57   \\ 
                            & LLaMA-3.1-8B-Instruct & 82.75   &83.91  & 54.16  & 50.57 \\
                            & LLaMA-3.2-3B-Instruct & 82.51   &82.99 & 49.18  & 48.83 \\
\midrule
\multirow{2}{*}{Others' approach}  & SpCQL  & 1.4  & 1.8  & 2.3  & 2.6  \\  
                            & Align-NL2GQL  & 82.99  & 84.06 &  54.21 &  52.86    \\
                            & $R^3$-NL2GQL     & 81.86  & 84.13 &  55.06 &  53.06 \\
\midrule
\multirow{1}{*}{Ours } 
        & Qwen-14B-Chat and ChatGPT-4o         & \textbf{85.44}  \textcolor{mylightred}{$\uparrow$2.45}    & \textbf{86.25}  \textcolor{mylightred}{$\uparrow$2.12} &\textbf{59.99}  \textcolor{mylightred}{$\uparrow$4.93} &\textbf{58.69}  \textcolor{mylightred}{$\uparrow$5.63} \\ 
\bottomrule
\end{tabular}
\caption{\label{table:main_result}
Comparison between our method and the baseline method. The bold numbers indicate the best results . The red upward arrow denotes an improvement, and the red number in parentheses indicates the exact improvement value compared to the best baseline method.
}
\end{table*}

\section{Experiment Results}
In this section, we evaluate the overall performance of NAT-NL2GQL. First, we introduce the experimental setup, followed by the presentation of the main results, an error analysis, further analysis, and an ablation study. Finally, we conclude with a case study to illustrate the effectiveness of our approach.

\subsection{Experimental Setup}
\noindent \textbf{Experiment Datasets.}
We conducted experiments on our constructed StockGQL dataset and the widely used SpCQL~\cite{guo2022spcql} dataset to evaluate the effectiveness of our approach. In particular, the SpCQL dataset uses GQL based on Cypher syntax rules, while the StockGQL dataset follows nGQL syntax rules. These differences, such as the use of "=" in Cypher versus "==" in nGQL, required us to make appropriate adjustments in the prompts during the experiments.

\noindent \textbf{Baseline Methods.}
To validate and compare the effectiveness of our method, as shown in Table ~\ref{table:main_result}, we selected three types of baseline methods: ICL approaches, fine-tuning approaches, and a method from previous related work. For the ICL approaches, the prompt format we designed is illustrated in Figure ~\ref{fig:icL}. In the fine-tuning approaches, the complete schema is incorporated into the input.

\begin{figure}[!htbp]
\centering
\resizebox{0.49\textwidth}{!}{
  \begin{tikzpicture}
    \node[draw, rectangle, inner sep=2mm, fill=mylightblue] (rect) {
      \begin{minipage}{\linewidth}
      \textbf{Instruction:}\\
You are an expert in NebulaGraph databases, please write the nGQL query corresponding to the given
Question directly based on the provided knowledge graph Schema and Examples. Ensure adherence to the nGQL conventions for naming
variables, entities, and attributes (e.g., ‘s.stock.name‘) and
verify that all conditional filters use ‘==‘ syntax, such as ‘s.stock.name == ’[s]’‘. Please provide the answer directly without any additional
explanation. Please provide the answer directly without any additional explanation. Please provide the
answer directly without any additional explanation.\\

\textbf{Output Format:} \\
Please output nGQL directly.\\

\textbf{Schema:}\\
\{SCHEMA\}\\

\textbf{Here are some examples:}\\
\{EXAMPLES\}\\
======== \textbf{Predict} ========\\
\textbf{Question:}\\
\{QUESTION\}\\
\textbf{nGQL:}\\
      \end{minipage}
    };
  \end{tikzpicture}
}
\caption{Prompt for In-Context Learning.}
\label{fig:icL}
\end{figure}

\noindent \textbf{Evaluation Metrics.}
We follow the approach in ~\cite{guo2022spcql,liang2024aligning}, using exact-set-match accuracy (\textbf{EM}) and execution accuracy (\textbf{EX}) to evaluate our method. EM measures the consistency of individual components, segmented by keywords, between the predicted query and its corresponding ground truth, while EX assesses the consistency of the execution results in the database.

\noindent \textbf{Implementation Details.}
Our experiments were performed on an A800 GPU. Taking into account hardware and time constraints, we selected GLM-4-9B-Chat, Qwen2.5-14B-Instruct, LLaMA-3.1-8B-Instruct, LLaMA-3.2-3B-Instruct, ChatGPT-3.5-Turbo, and ChatGPT-4o as the LLMs backbone. The number of demonstrations $k$ was uniformly set to $4$ in all experiments.
The threshold \(\mathcal{\gamma}\) of LSH is set to $0.6$.
\subsection{Main Results}
An analysis of the main experimental results in Table~\ref{table:main_result} leads to the following conclusions:

First, our approach demonstrates a significant improvement over all baseline methods. Specifically, on the StockGQL dataset, it outperforms the best baseline method by 2.45\% on the EM metric and by 2.12\% on the EX metric. Similarly, on the SpCQL dataset, it surpasses the best baseline by 4.93\% on the EM metric and by 5.63\% on the EX metric.

Second, the performance of the ICL method is relatively poor for the NL2GQL task, likely due to the general lack of high-quality GQL-related corpora during the training of the base models. To improve its performance, one possible approach would be to collect high-quality GQL corpora and continue training the base LLMs.

Third, it is evident that the SpCQL dataset is more challenging than the StockGQL dataset, primarily because SpCQL questions contain more entities with names that differ from those in the database. This also explains why our method achieves greater improvement on the SpCQL dataset, as we have developed an entity alignment approach.

Interestingly, by comparing the performance of LLaMA-3.1-8B-Instruct and LLaMA-3.2-3B-Instruct, we found that the former performs significantly better than the latter with the ICL method. However, after fine-tuning, their performances are nearly identical. This suggests that smaller models may not be well-suited for the ICL approach but are more suitable for fine-tuning when sufficient data is available.

\subsection{Error Analysis}
\begin{figure}[]
\centering
\includegraphics[width=0.48\textwidth]{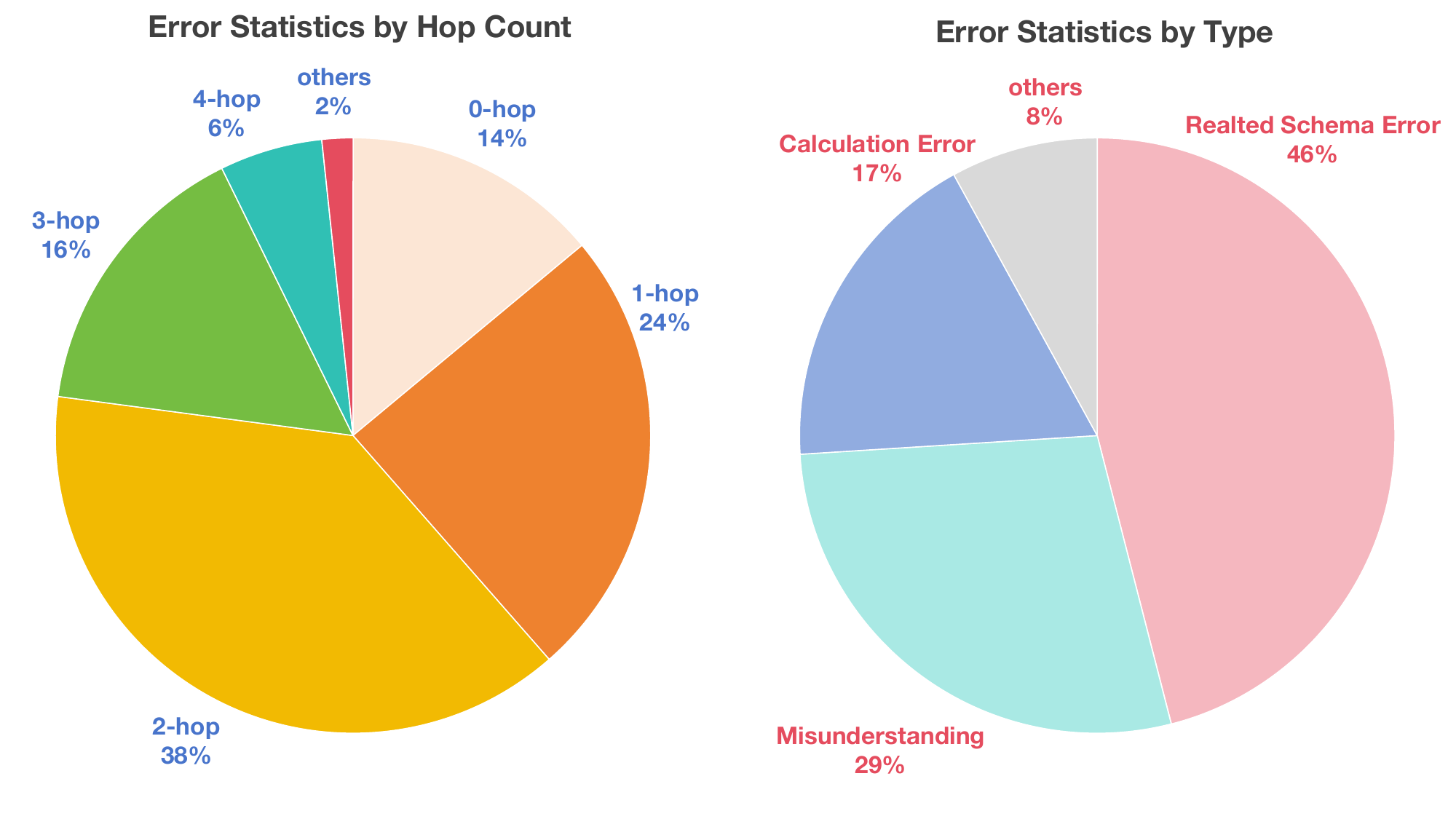}
\caption{Error analysis statistics chart. On the left is the error statistics based on the hop count of the jump in the NL, and on the right is the error statistics based on the error type.  
}
\label{fig:error_ana}
\end{figure}

To further evaluate the performance of our method, we analyzed the errors on StockGQL, categorizing them by hop count and error type. As shown in Figure ~\ref{fig:error_ana}, the "Error Statistics by Hop Count" segment reveal that the majority of errors occur in the 0 to 3 hop range, which corresponds to the higher proportion of test data in this range. The "Error Statistics by Type" reveal that nearly half of the errors are related to schema extraction. Therefore, improving the accuracy of related schema extraction would be a promising strategy. Furthermore, a significant portion of the errors arises from misunderstandings of the question, highlighting that question comprehension remains a critical challenge, especially for more colloquial or ambiguous queries.

\subsection{Further Analysis}
\begin{figure}[]
\centering
\includegraphics[width=0.48\textwidth]{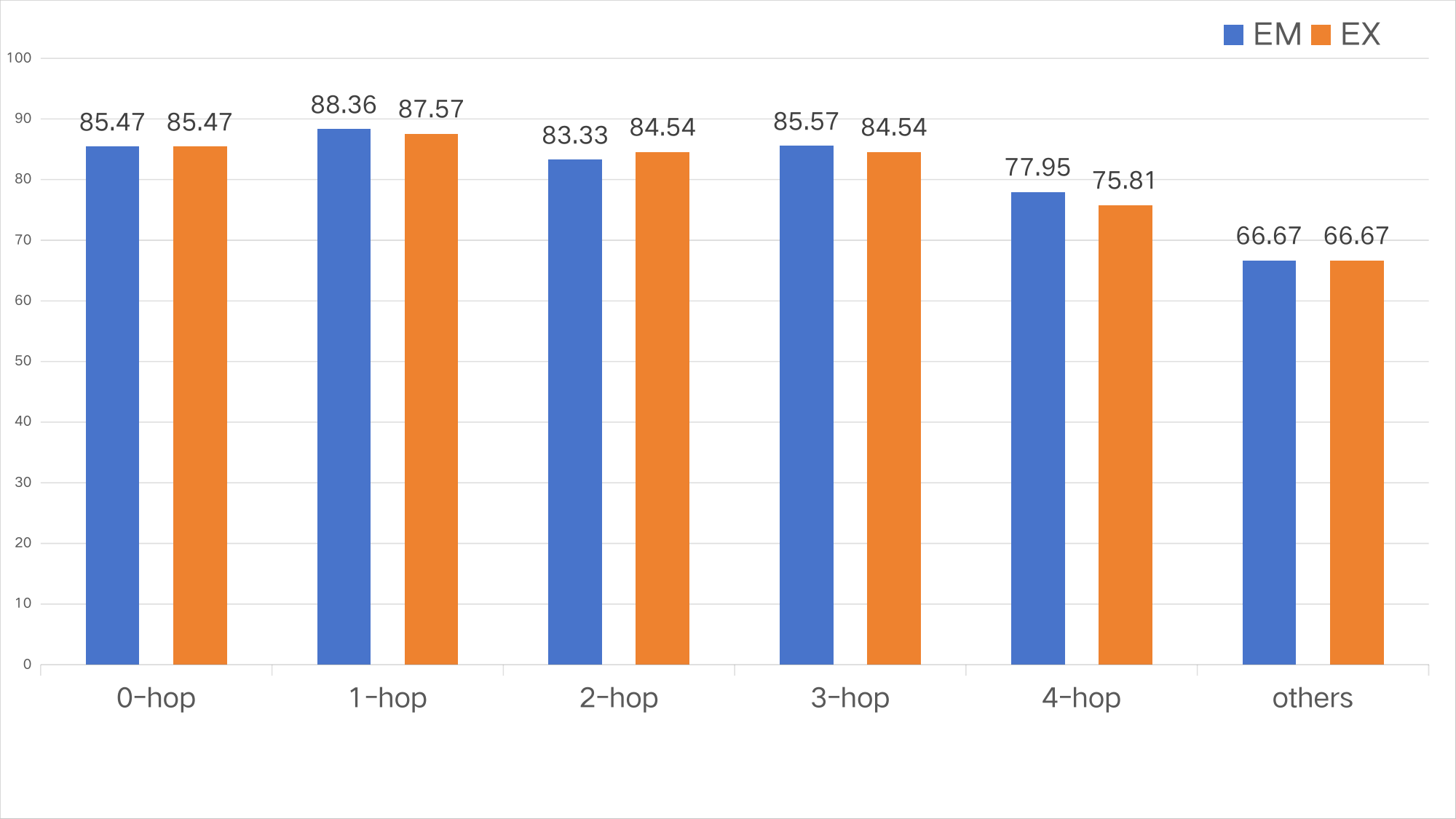}
\caption{The accuracy of the EM and EX metrics of our method on StockGQL, statistically based on hop count.}
\label{fig:breakdown_result}
\end{figure}

\noindent \textbf{Breakdown Analysis.} 
To further analyze the model's performance on queries of varying difficulty, we calculated the accuracy of our method on StockGQL, segmented by hop for each question type. The results in Figure~\ref{fig:breakdown_result} show that, except for 0-hop queries, the general trend is a decrease in accuracy as the hop number increases, indicating higher query difficulty. The lower accuracy for 0-hop questions compared to 1-hop is due to their more flexible, conversational nature. Additionally, as the number of hops increases, it becomes more challenging for the model to correctly generate the corresponding GQL, reflecting the growing complexity of processing multi-hop queries. These queries require the model to navigate more intricate dependencies between different parts of the query. Furthermore, this highlights the need for advanced techniques to effectively handle such complex queries, particularly those involving multiple relational dependencies across different entities.

\begin{table}[]
\centering
\small
\begin{tabular}{ll}
\toprule
\textbf{Dataset)} & \textbf{Acc(\%)}  \\
\midrule
Ours  &  \bf{86.00} \\
Ours(w/o filtering)    &  72.50 \\
Align-NL2GQL      & 80.15  \\
$R^3$-NL2GQL      & 81.04 \\
\bottomrule
\end{tabular}
\caption{\label{related_schema_acc}
Comparison of accuracy across different methods for extracting related schemas on StockGQL.
}
\end{table}

\noindent \textbf{Accuracy of Related Schema Extraction.} 
Existing works~\cite{liang2024aligning,zhou2024r} and Table~\ref{schema_ana} have emphasized the importance of related schema extraction. Therefore, we compared the accuracy of our method with $R^{3}$-NL2GQL and Align-NL2GQL on the StockGQL dataset. The experimental results, presented in Table~\ref{related_schema_acc}, show that our method achieves the highest accuracy. Although our method already demonstrates high accuracy, there is still considerable room for improvement. Enhancing the accuracy of the extracted related schema remains a promising direction for future exploration.

\noindent \textbf{Impact of the Related Schema.}
We have consistently emphasized the crucial role of the related schema in the accuracy of the generated GQL. To assess the precise impact of the related schema on GQL accuracy, we conducted three experiments using StockGQL. The first two experiments focused on fine-tuning Qwen2.5-14B-Instruct with LoRA~\cite{hu2021lora}, utilizing the related schema corresponding to the GQL label in the training set. In the "Golden Related Schema" experiment, we used the related schema from the test dataset that matched the GQL label as input. In contrast, the "Error Related Schema" experiment involved modifying this schema before inputting it. Finally, in the "All Schema" experiment, the full schema was employed for both training and testing. The results presented in Table~\ref{schema_ana} show that when an erroneous related schema is provided, there is a high likelihood of generating incorrect GQL. Furthermore, using the correct related schema significantly improves performance compared to using the full schema. This further emphasizes the critical role of related schema accuracy in determining the precision of the generated GQL.

\begin{table}[ht]
\centering
\small
\begin{tabular}{lll}
\toprule
\textbf{Dataset} & \textbf{EM(\%)} & \textbf{EX(\%)} \\
\midrule
Error Related Schema    &  20.34     &  21.16 \\
All Schema      & 82.67 & 84.05 \\
Golden Related Schema  &  91.46    &  92.84 \\
\bottomrule
\end{tabular}
\caption{\label{schema_ana}
The table presents the results showing the impact of the related schema on the accuracy of the generated GQL for StockGQL.
}
\end{table}

\noindent \textbf{Performance with Various  Base LLMs.} In our approach, the Preprocessor and Refiner agents use ChatGPT-4o, while the Generator is fine-tuned using LoRA on Qwen2.5-14B-Instruct. To investigate the impact of base LLMs on each component agent, we conducted experiments using several base LLMs for each agent and compared the results. The experimental results are summarized in Table~\ref{table:basellm}. These findings indicate that the Generator agent demonstrates greater robustness to the choice of base LLMs after fine-tuning. In contrast, the Preprocessor agent and Refiner agent, which utilize unmodified base LLMs, exhibit higher sensitivity to the underlying model, leading to a significant impact on the overall system performance.

\begin{table}[ht!]
\centering
\small
\renewcommand\arraystretch{0.8}
\begin{tabular}{l l c c }
\toprule
\textbf{Agent} & \textbf{LLM} & \textbf{EM(\%)} & \textbf{EX(\%)} \\
\midrule
\multirow{4}{*}{Preprocessor}
                              
                              & Qwen2.5-14B-Instruct & 77.95   & 79.01  \\
                              & ChatGPT-3.5-Turbo    & 80.88 & 79.98   \\
                              & \textbf{ChatGPT-4o}   & \textbf{85.44}  & \textbf{86.25 }   \\
\midrule
\multirow{2}{*}{Generator}  &GLM-4-9B-Chat   
                             & 85.03 & 85.84   \\
                            & LLaMA-3.1-8B-Instruct & 85.35   &86.09   \\
                            & LLaMA-3.2-3B-Instruct & 85.19   &85.92 \\
                            & \textbf{Qwen2.5-14B-Instruct} & \textbf{85.44}  & \textbf{86.25 }   \\ 
\midrule
\multirow{2}{*}{Refiner}   & Qwen2.5-14B-Instruct &
                              84.21   & 84.95  \\
                              & ChatGPT-3.5-Turbo    & 84.87 & 85.68   \\
                              & \textbf{ChatGPT-4o}  & \textbf{85.44}  & \textbf{86.25 }  \\
\bottomrule
\end{tabular}
\caption{\label{table:basellm}
Compare the impact of different base LLMs on the final performance of NAT-NL2GQL on StockGQL. The base LLMs used in our method are highlighted.
}
\end{table}

\subsection{Ablation Study}
We conducted an ablation study to evaluate the effectiveness of the components in our method, with the results presented in Table~\ref{table:ablation}. As shown in the table, removing any component results in a performance decline, highlighting the crucial role of each component in our method. Notably, replacing the fine-tuned generator with the ICL method of ChatGPT-4o causes the most significant performance drop, which aligns with the findings in ~\ref{schema_ana}. Additionally, the "Without Regeneration" setting demonstrates that the Refiner detects related schema extraction errors and provides the Preprocessor with feedback to re-extract, restarting the iterative process. While this part yields the least improvement, it still contributes to a noticeable enhancement.

\begin{table}[]
\centering
\small
\renewcommand\arraystretch{0.8}
\begin{tabular}{lcc}
\toprule
\multicolumn{1}{l}{\textbf{Method}} &     \textbf{EM(\%)}           & \textbf{EX(\%)}          \\ 
\midrule
Ours                     & \textbf{85.44}        & \textbf{86.25}       \\ 
  \midrule
   Without Preprocessor     & 80.55 \textcolor{mylightgreen}{$\downarrow${(4.88)}} &  80.06 \textcolor{mylightgreen}{$\downarrow${(6.19)} }    \\
   Generator -> ChatGPT-4o      &  50.04 \textcolor{mylightgreen}{$\downarrow${(35.40)}}      & 47.93  \textcolor{mylightgreen}{$\downarrow${(38.32)} }    \\
   Without  Refiner  & 83.40 \textcolor{mylightgreen}{$\downarrow${(2.04)}} & 83.16  \textcolor{mylightgreen}{$\downarrow${(3.09)}}      \\ 
   Without Regeneration   & 84.21  \textcolor{mylightgreen}{$\downarrow${(1.23)}}      & 84.13  \textcolor{mylightgreen}{$\downarrow${(2.12)} }    \\
\bottomrule
\end{tabular}
\caption{\label{table:ablation}
 Ablation study of our method on StockGQL. The green downward arrow denotes a decrease, and the green number in parentheses indicates the precise decrease value.}
\end{table}


\subsection{Case Study}

\begin{table*}[ht!]
\centering
\small
\renewcommand\arraystretch{0.8}
\begin{tabular}{lll}
    \toprule
    \textbf{NL} &
    \multicolumn{2}{l}{\makecell[c]{\begin{CJK*}{UTF8}{gbsn}持有华强股票且持仓比例超过7\%的基金经理？ \end{CJK*} \\
    (The fund manager who manages the fund that holds Huaqiang stock with a holding ratio exceeding 7\%?)}} \\
    \midrule
    \textbf{Formal expression} &
    \multicolumn{2}{l}{\makecell[c]{\begin{CJK*}{UTF8}{gbsn}持有华强科技股票且持仓比例超过7\%的基金由哪位基金经理管理？ \end{CJK*} \\
    (Which fund manager manages the fund that holds \\Huaqiang Technology stock with a holding ratio exceeding 7\%?)}} \\
    \midrule
    \textbf{Method} &  \textbf{Related Nodes and Edges}  & \textbf{Output} \\
    \midrule
    \makecell[l]{ICL(ChatGPT-4o )} & 
    \makecell[l]{
    full schema
    }&
    \makecell[l]{\color{mylightred}{MATCH (s:stock\{name: '\begin{CJK*}{UTF8}{gbsn}华强\end{CJK*}'\})<-[h:manage]-(fm:fund\_manager) }\\
    \color{mylightred}{WHERE h.position\_ratio > 7\% 
    RETURN fm.name}}  \\
    \midrule
    \makecell[l]{Fine-Tuning(full schema) \\Qwen2.5-14B-Instruct} &
    \makecell[l]{
    full schema
    }&
   \makecell[l]{MATCH (s:stock\{name: '\begin{CJK*}{UTF8}{gbsn}\color{mylightred}{华强}\end{CJK*}'\})<-[h:hold]-\\
    (pof:fund)<-[:manage]-(fm:fund\_manager) \\
    \color{mylightred}{WHERE h.hold.position\_ratio > 7\% }\\
    RETURN fm.fund\_manager.name}  \\
    \midrule
    \makecell[l]{Align-NL2GQL} 
    & \makecell[l]{
    \color{darkblue}{Nodes:[fund\_manager,fund,stock]}\\ 
    \color{darkblue}{Edges: [manage,hold]}
    }
    & \makecell[l]{MATCH (s:stock\{name: '\begin{CJK*}{UTF8}{gbsn}\color{mylightred}{华强}\end{CJK*}'\})<-[h:hold]-\\
    (pof:fund)<-[:manage]-(fm:fund\_manager) \\
    WHERE h.position\_ratio > 7\% 
    RETURN fm.fund\_manager.name}  \\
    \midrule
    \makecell[l]{$R^3$-NL2GQL}
    & \makecell[l]{
    \color{mylightred}{Nodes:[fund\_manager,stock]}\\ 
    \color{mylightred}{Edges: [hold]}
    }
    &  \makecell[l]{\color{mylightred}{MATCH (s:stock\{name: '\begin{CJK*}{UTF8}{gbsn}华强科技\end{CJK*}'\})<-[h:hold]-(fm:fund\_manager) }\\
    WHERE h.position\_ratio > 7\% 
    RETURN fm.fund\_manager.name}    \\
    \midrule
    \makecell[l]{Ours} 
    &   \makecell[l]{
    \color{darkblue}{Nodes:[fund\_manager,fund,stock]}\\ 
    \color{darkblue}{Edges: [manage,hold]}
    }
    & \makecell[l]{\color{darkblue}{MATCH (s:stock\{name: '\begin{CJK*}{UTF8}{gbsn}华强科技\end{CJK*}'\})<-[h:hold]-}\\
    \color{darkblue}{(pof:fund)<-[:manage]-(fm:fund\_manager) }\\
    \color{darkblue}{WHERE h.position\_ratio > 7\% RETURN fm.fund\_manager.name}} \\
    \bottomrule
\end{tabular}
\caption{A case study in the StockGQL dataset is presented, displaying the results of both our method and the baseline methods. Due to space limitations, the table uses "Related Nodes and Edges" rather than listing the full details of the related schema. The segments with predicted errors are highlighted in \textcolor{mylightred}{red}, while the correct ones are marked in \textcolor{darkblue}{blue}. }
\label{case_study}
\end{table*}
To further demonstrate the strengths of our method, we present a detailed case study in Table ~\ref{case_study}. From the case, we observe that baseline methods either extract the wrong related schema, generate GQL with syntax errors, or fail to recognize colloquial variations of named entities. In contrast, our approach accurately extracts the related schema, even for multi-hop queries, and effectively interprets colloquial variations of named entities. This ensures that entity names are recognized and accurately reflected in the generated GQL, even when the input deviates from standard formal representations. This highlights the robustness and adaptability of our method in handling complex and varied queries, further reinforcing its effectiveness in real-world applications.

\section{Related works}
\subsection{NL2GQL}
NL2GQL is a typical NLP task that has emerged with the widespread adoption of graph data and can be classified as a seq-to-seq task~\cite{guo2022spcql, zhao2023cyspider}. Its primary function is to convert users' NL questions into GQL queries that can be executed on a graph database. This task involves  user queries understanding,   graph schema linking, and GQL generating ~\cite{liang2024aligning, zhou2024r}. Early efforts focused on using hand-crafted rules to translate NL into GQL~\cite{zhao2022natural}. Modern approaches primarily incorporate state-of-the-art (SOTA) models to optimize performance. We categorize LLM-based NL2GQL methods into two types: PLMs-based 
 methods and LLMs-based Methods.

\noindent \textbf{PLMs-based methods}.
Fine-tuning PLMs within a sequence-to-sequence framework is one of the most widely used approaches for generative tasks in NLP.
Initially, ~\cite{guo2022spcql} constructed a text-to-Cypher dataset and designed three baselines: seq2seq, seq2seq + attention~\cite{dong2016language}, and seq2seq + copying~\cite{gu2016incorporating}. However, the results on the two evaluation metrics, EX and EX, were not satisfactory. 
Reference ~\cite{tran2024robust} employs the BERT~\cite{kenton2019bert} model for key-value extraction and uses GraphSAGE~\cite{hamilton2017inductive} to analyze the relational properties of the database. These features are then fed into a transformer to generate the Cypher query. Their proposed small Text-to-Cypher dataset outperforms seq2seq models like T5~\cite{raffel2020exploring} and GPT-2~\cite{radford2019language}. 
Reference ~\cite{liang2024kei} introduces the KEI-CQL framework, a heuristic-like approach that utilizes pre-trained language models to extract semantic features from natural language queries and populate predefined slots in Cypher query sketches, effectively addressing the NL2GQL challenge.

\noindent \textbf{LLM-based  methods}.
Leveraging the powerful understanding and generation capabilities of LLMs to tackle the NL2GQL task has become a recent research hotspot. 
Reference ~\cite{tao2024finqa} attempts to combine heuristic methods with LLM-based approaches. They first extract GQL clauses using heuristic rules, then concatenate these clauses to form a complete GQL, and finally use an LLM for refinement. 
Reference ~\cite{zhou2024r} deconstructs the NL2GQL task into individual subtasks, using a combination of smaller models and LLMs for each stage. Specifically, smaller models are employed during the initial ranking and rewriting phases, while an LLM is used for the final generation step. 
In contrast, ~\cite{liang2024aligning} aligns LLMs with domain-specific graph databases to address NL2GQL tasks within those databases. They first construct an NL2GQL dataset based on a domain-specific graph database, then fine-tune an LLM with this dataset, enabling the LLM to effectively tackle NL2GQL tasks in the specific domain. 
However, streaming-based task decomposition methods often struggle with error accumulation. In response to the observed challenge, we introduce the NAT-NL2GQL framework.
\subsection{Text2SQL}

Text2SQL is a task in NLP that is quite similar to NL2GQL, as both involve transforming user queries into statements that can be executed on a database. 
Recently, there have been many efforts to apply LLMs to solve Text2SQL, and these methods have achieved good results~\cite{pourreza2024chasesq,maamari2024death,li2024can,caferouglu2024sql}.
However, there are significant differences between the two. 
    The diversity inherent in GQL presents a series of challenges. Unlike SQL, which has a well-established and standardized query language for relational databases, GQL lacks a unified standard~\cite{zhou2024r}. This deficiency creates obstacles in various areas, including dataset construction, the development of models capable of generalizing across different databases, and the establishment of consistent training paradigms.
    There is a difference in query objectives. NL2GQL aims to execute queries on graph databases, whereas Text2SQL targets relational databases. Graph databases feature more flexible data structures and complex relationships, requiring NL2GQL to manage a wider variety of queries and data relationships~\cite{liang2024aligning}.
    The flexibility of query languages differs. GQL is more flexible compared to SQL, allowing for complex queries on nodes and edges in a graph database, while SQL is constrained by the fixed structure and syntax of relational databases.
    The complexity of query paths is notable. Queries in graph databases often involve intricate paths between multiple nodes and edges. NL2GQL must handle these complex paths and translate natural language questions into corresponding GQL queries, adding to the overall complexity of the task.
    There is a greater variety of keyword types in GQL compared to SQL~\cite{guo2022spcql}. GQL encompasses more keyword types, reflecting the diverse data structures and query requirements in graph databases. NL2GQL must recognize and process these different types of keywords, further complicating the task.
In summary, NL2GQL is more complex than Text2SQL due to its handling of graph database queries, the flexibility of GQL, the complexity of data paths and the variety of keyword types. Given these differences, it is challenging to directly transplant methods from the Text2SQL task to the NL2GQL task.

\subsection{KBQA}
Knowledge-Based Question Answering (KBQA) systems utilize structured knowledge bases (KBs) to answer user queries. SP-based methods, commonly referred to as NL2SPARQL, first transform natural language questions into SPARQL queries, which are then executed on the KB to retrieve results~\cite{lan2021survey}. This approach shares similarities with NL2GQL; however, a significant difference between NL2GQL and NL2SPARQL in the KGQA domain lies in the complexity of data storage and query languages. Graph databases, which manage data with intricate relationships, introduce additional complexity~\cite{liang2024aligning}. 
Furthermore, NL2GQL demands greater attention to schema information, as entities in graph databases can encompass a diverse range of attribute types~\cite{zhou2024r}. NL2GQL is also characterized by complex graph modalities, a wide variety of query types, and the unique nature of GQLs~\cite{zhou2024r}. As a result, directly applying KBQA methods to the NL2GQL task is impractical.

\section{Conclusion and future work}
\label{sec:conc}

In this paper, we propose the NAT-NL2GQL framework to tackle the NL2GQL task. Specifically, our framework consists of three synergistic agents: the Preprocessor Agent, the Generator Agent, and the Refiner Agent. Additionally, we have constructed the StockGQL dataset based on a graph database for NL2GQL. Experimental results demonstrate that our approach significantly outperforms baseline methods.

One promising research direction is to design more innovative methods for extracting the related schema for a query, such as leveraging advanced graph neural networks, utilizing context-aware techniques, or incorporating adaptive learning mechanisms to dynamically refine the extraction process based on query complexity and context. Another direction is to increase the context length of LLMs and enhance their ability to understand the schema of a graph database, enabling them to process larger, more complex queries and better capture the relationships between entities within the schema for more accurate GQL generation, without the need for extracting the related schema. 

\bibliographystyle{IEEEtran}
\balance
\bibliography{ref.bib}


\end{document}